\pdfoutput=1

\documentclass[11pt]{article}

\usepackage[]{ACL2023}

\usepackage{times}
\usepackage{latexsym}

\usepackage[T1]{fontenc}

\usepackage[utf8]{inputenc}

\usepackage{microtype}

\usepackage{inconsolata}

\newcommand{\quotes}[1]{``#1''}
\usepackage{amsmath}
\usepackage{tikz}
\def\checkmark{\tikz\fill[scale=0.4](0,.35) -- (.25,0) -- (1,.7) -- (.25,.15) -- cycle;} 
\usepackage[export]{adjustbox}
\usepackage{booktabs}
\usepackage{arydshln}
\usepackage{multirow}
\usepackage{hhline}
\usepackage{xcolor}

\newcommand\blfootnote[1]{
    \begingroup
    \renewcommand\thefootnote{}\footnote{#1}
    \addtocounter{footnote}{-1}
    \endgroup
}

\usepackage{hyperref}

\usepackage{amssymb}
\usepackage{pifont}
\newcommand{\xmark}{\ding{55}}%

\title{InstructCMP: Length Control in Sentence Compression through Instruction-based Large Language Models}

\author{Juseon-Do$^1$, $^*$Jingun Kwon$^1$, Hidetaka Kamigaito$^2$, and Manabu Okumura$^3$ \\
 $^1$Chungnam National University, $^2$Nara Institute of Science and Technology (NAIST) \\
 $^3$Tokyo Institute of Technology \\
 {\tt doju00@o.cnu.ac.kr} \\
 {\tt jingun.kwon@cnu.ac.kr} \\
 {\tt kamigaito.h@is.naist.jp} \\
 {\tt oku@pi.titech.ac.jp}
 \\}

\begin{document}
\maketitle
\begin{abstract}
Extractive summarization can produce faithful summaries but often requires additional constraints such as a desired summary length. Traditional sentence compression models do not typically consider the constraints because of their restricted model abilities, which require model modifications for coping with them. To bridge this gap, we propose Instruction-based Compression (InstructCMP), an approach to the sentence compression task that can consider the length constraint through instructions by leveraging the zero-shot task-solving abilities of Large Language Models (LLMs). For this purpose, we created new evaluation datasets by transforming traditional sentence compression datasets into an instruction format. By using the datasets, we first reveal that the current LLMs still face challenges in accurately controlling the length for a compressed text. To address this issue, we propose an approach named \quotes{length priming,} that incorporates additional length information into the instructions without external resources. While the length priming effectively works in a zero-shot setting, a training dataset with the instructions would further improve the ability of length control. Thus, we additionally created a training dataset in an instruction format to fine-tune the model on it.
Experimental results and analysis show that applying the length priming significantly improves performances of InstructCMP in both zero-shot and fine-tuning settings without the need of any model modifications.\blfootnote{$^*$ corresponding author}
\end{abstract}

\section{Introduction}
Sentence compression is a task of creating a concise summary from an original sentence while conveying its key information, by deleting words in the sentence. Generally, sentence compression in extractive summarization 
provides 
more faithful summaries than abstractive summarization~\cite{Cao2018FaithfulTT}.

While traditional sentence compression methods used tree trimming, the approaches can be affected by parsing errors~\cite{jing-2000-sentence,knight,berg-kirkpatrick-etal-2011-jointly,filippova-altun-2013-overcoming}. The introduction of LSTM-based Seq2Seq approaches aims to address this issue although their performance tends to degrade in handling longer sentences~\cite{filippova-etal-2015-sentence}.
To solve this problem, \newcite{Kamigaito_Okumura_2020} incorporated syntactic dependency trees into the Seq2Seq attention mechanism~\cite{kamigaito-etal-2018-higher} by jointly learning the dependency trees and sentence compression models. However, the state-of-the-art model required a considerable amount of ground-truth data for training~\cite{filippova-altun-2013-overcoming,hasegawa-etal-2017-japanese}. 

\begin{figure}[t!]
\centering
\includegraphics[width=0.9\columnwidth]{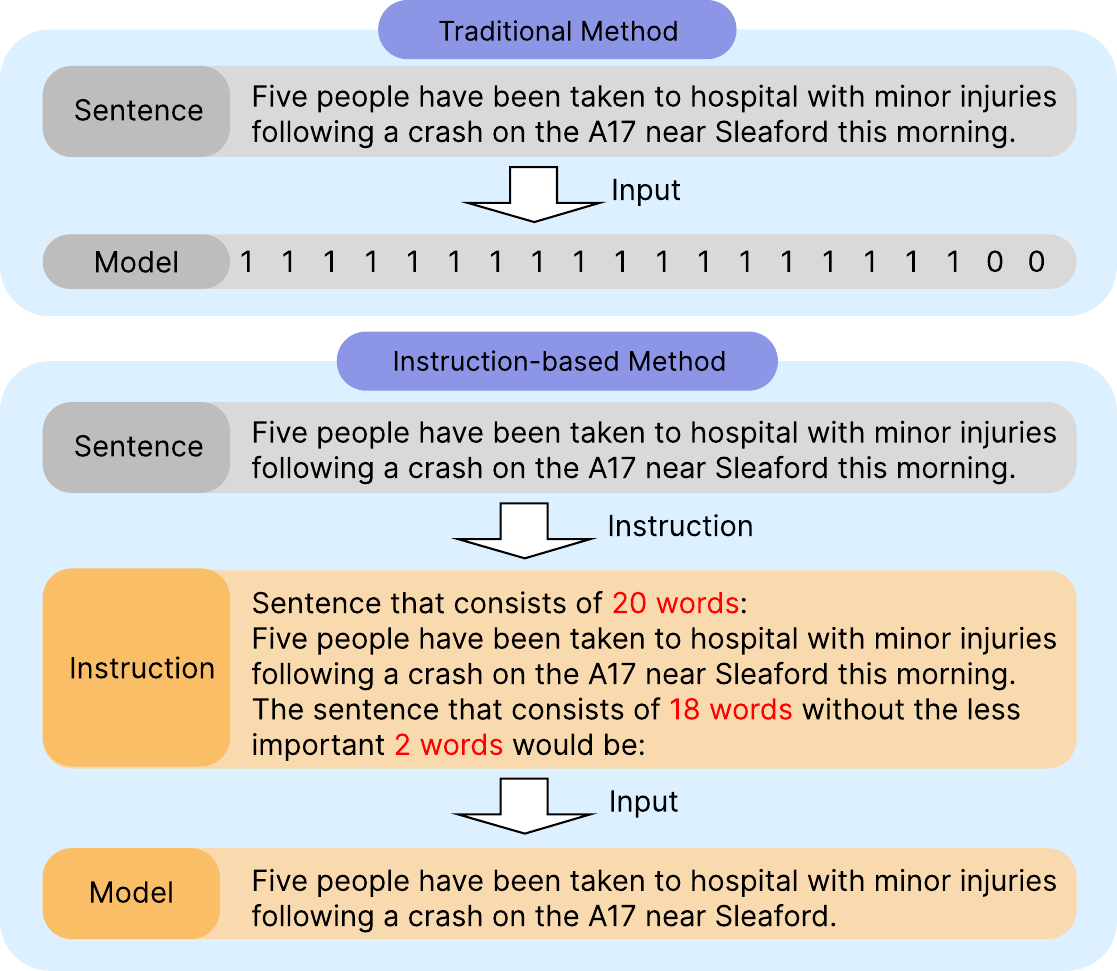}
\caption{Process of transforming a traditional labeled dataset into an instruction-based format. The binary output of \quotes{1} or \quotes{0} from the traditional methods corresponds to keeping or dropping words, respectively. Length constraints in \quotes{length priming} are highlighted in \textcolor{red}{red} in the instruction.}
\label{fig:figure1}
\end{figure}

Recently, unsupervised sentence compression has gained attention by exploiting BERT-based encoder models~\cite{devlin-etal-2019-bert}. These models incorporated various scoring functions that target improving fluency and faithfulness in compression without relying on ground-truth data~\cite{Deleter,zhou-rush-2019-simple,schumann-etal-2020-discrete,ghalandari-etal-2022-efficient}. However, these approaches are inefficient because they require extensive model modifications, such as including classifiers or modifying beam search for objective-specific fine-tuning. 

In general, summarization requires additional constraints such as a summary length~\cite{takase-okazaki-2019-positional,dou-etal-2021-gsum,kwon-etal-2023-abstractive}. The traditional task setting for sentence compression often did not consider this factor because of the restricted model abilities, which require model modifications to handle such constraints~\cite{schumann-etal-2020-discrete,ghalandari-etal-2022-efficient}. 

Recently, LLMs have gained considerable attention for their remarkable zero-shot task-solving abilities, especially under instruction-based settings~\cite{neuips2022,weifinetuned}. 
Inspired by these latest advancements, we present Instruction-based Compression (InstructCMP), a novel approach to sentence compression that accommodates 
a length constraint through explicit instructions, without necessitating model modifications. 
To the best of our knowledge, this approach represents the first 
implementation of 
sentence compression in an instruction-based framework. For this purpose, we transformed traditional sentence compression datasets into an instruction-based format for evaluation. 

However, recent LLMs do not consistently generate an output of the precise length, even when specific instructions to include such constraints are provided in a zero-shot manner~\cite{pmlr-v202-zhou23g,qin-etal-2023-chatgpt}. Furthermore, as we validate it later, even when testing with the latest powerful models, such as ChatGPT (GPT-4) and ChatGPT (GPT-4-1106-preview),\footnote{\url{https://chat.openai.com/}} accurately adhering to length constraints remains a substantial challenge.

To address this problem, we propose an instruction approach for better length control, which is named \quotes{length priming.} We incorporate additional length information~\cite{misra-etal-2020-exploring} into the instruction. In addition to 
specifying the number of deleted words for the desired length, we include the length to be retained and the number of words in the source sentence in the instruction, 
without any external resources. To further improve length controllability, we additionally created a training dataset with the instructions to fine-tune the model using the dataset.
Figure~\ref{fig:figure1} shows the transformation process for an instruction format. 

We conducted experiments on four benchmark datasets and performed an in-depth analysis to evaluate the effectiveness of LLMs in compressing sentences under the length constraint. The analysis considers the following factors: the model type and the number of parameters for 
the model size. 
Experimental results show that InstructCMP with length priming compresses sentences in a zero-shot setting while successfully keeping the desired length without model modifications. The performance can be further improved by fine-tuning it with the created instruction-based training dataset. The \quotes{length priming} method proves effective in both zero-shot and fine-tuning settings, as shown by significant improvements in the ROUGE metrics and adherence to the length constraint, even when using ChatGPT (GPT-4) and ChatGPT (GPT4-1106-preview). Our in-depth analysis also showed that InstructCMP can compress sentences while maintaining faithfulness. 
Our experiments show 
that instruction-based models like ChatGPT can effectively control the length when provided with more specific length-related information.\footnote{Our code and datasets are available at:  \url{https://github.com/JuseonDo}.}

\begin{table}[t]
\begin{adjustbox}{width=1\columnwidth,center}
\centering

\begin{tabular}{lcc}
\toprule
\textbf{Work} & \textbf{Length Const.} & \textbf{Mod.}\\
\midrule
\newcite{filippova-etal-2015-sentence}$^{\ast}$  & \xmark & \xmark \\
\newcite{zhao-etal-2018-language}$^{\ast}$  & \xmark & \xmark \\
\newcite{Kamigaito_Okumura_2020}$^{\ast}$  & \xmark & \xmark \\
\newcite{schumann-etal-2020-discrete}  & \checkmark & \xmark \\
\newcite{ghalandari-etal-2022-efficient} & \checkmark & \xmark \\
\midrule
Ours (InstructCMP)  & \checkmark &  \checkmark  \\
\bottomrule

\end{tabular}
\end{adjustbox}
\caption{Comparison of various sentence compression models with InstructCMP. $\ast$ indicates that the model was learned in a supervised manner, while others were learned in an unsupervised manner. Mod. indicates a requirement of model modifications for constraints. }
\label{tab:previousworks}
\end{table}

\section{Problem Statement}
The traditional approach to sentence compression is considered as a sequential labeling task~\cite{filippova-etal-2015-sentence,wang-etal-2017-syntax,zhao-etal-2018-language,Kamigaito_Okumura_2020,schumann-etal-2020-discrete,ghalandari-etal-2022-efficient}. Each source token in a sequence, represented as $\textbf{x}=\{x_0, x_1, ..., x_n\}$, is processed using a sentence compression model to predict a corresponding label sequence, which is $\textbf{y}=\{y_0, y_1, ..., y_n\}$, where $y_i \in \{1, 0\}$.

\begin{table*}[t]
\begin{adjustbox}{width=1.8\columnwidth,center}
\centering
\small
\renewcommand{\arraystretch}{1.2}
\begin{tabular}{ccl}
\toprule
\# &\textbf{Constraint} & \textbf{Instruction}\\
\midrule
1 & \xmark & Sentence:\textbackslash n\{src\}\textbackslash nThe sentence without the less important words would be:\textbackslash n\\
\midrule
2 & Length w/o priming & Sentence:\textbackslash n\{src\}\textbackslash nThe sentence without the less important \{del\} words would be:\textbackslash n\\
\midrule
\multirow{2}{*}{3} & \multirow{2}{*}{Length}  & Sentence that consists of \{src len\} words:\textbackslash n\{src\}\textbackslash nThe sentence that consists of \{keep\} words \\
&  & without the less important \{del\} words would be:\textbackslash n \\
\cmidrule{2-3}
\multirow{2}{*}{3-1} & Length & Sentence that consists of \{src len\} words:\textbackslash n\{src\}\textbackslash nThe sentence without the less important \{del\} \\
& w/o tgt priming & words would be:\textbackslash n\\
\cmidrule{2-3}
\multirow{2}{*}{3-2} & Length  & Sentence:\textbackslash n\{src\}\textbackslash nThe sentence that consists of \{keep\} words without the less important  \{del\} \\
& w/o src priming &  words would be:\textbackslash n \\

\bottomrule
\end{tabular}
\end{adjustbox}
\caption{Instruction formats for length constraints, created by transforming a traditional dataset. \quotes{src} indicates the placeholder for a source sentence. 
\quotes{del} denotes the placeholder for 
the number of deleted words. \quotes{keep} and \quotes{src len} denote additional length information.}
\label{tab:instruction_format}
\end{table*}

While the method is straightforward, it has limitations in incorporating additional constraints such as 
a desired length. Addressing these requirements in the traditional approach typically involves 
modifications to the model, 
which is inefficient~\cite{schumann-etal-2020-discrete,ghalandari-etal-2022-efficient}.

To overcome these limitations, we utilize the recent powerful instruction-based LLMs for the sentence compression task~\cite{touvron2023llama,chung2022scaling}. Table~\ref{tab:previousworks} shows a comparison between previous work on traditional sentence compression and InstructCMP. Unlike the previous work, InstructCMP incorporates a length constraint directly into the instruction format, allowing models to process and learn the constraint as a part of their input. This allows an efficient and flexible solution for practical sentence compression, without extensive model modifications. 

\section{Instruction-based Compression}
In this section, we describe 
InstructCMP. We consider \quotes{length priming} for a length constraint in it. We created new evaluation datasets by transforming traditional sentence compression datasets into an instruction format. To further improve performances of 
InstructCMP, we also created a new training dataset in an instruction-based template. 

\subsection{Instruction Template}
Table~\ref{tab:instruction_format} shows instructions that include a length constraint. The first instruction permits InstructCMP to compress text by deleting words without any constraints. However, in general, summarization requires 
a desired length for compressed texts~\cite{makino-etal-2019-global,
dou-etal-2021-gsum,he-etal-2022-ctrlsum,kwon-etal-2023-abstractive}.


\noindent \textbf{Length Priming.} To apply the length constraint, we first construct an instruction that deletes words to meet a desired length 
(Constraint 2). 
It is easy to calculate 
the number of words to be deleted for any desired length.

However, LLMs do not consistently follow instructions, particularly when processing length constraints~\cite{pmlr-v202-zhou23g,qin-etal-2023-chatgpt}. To address this issue, we propose the \quotes{length priming} method for the length constraint instruction for enhanced length comprehension. Constraint 3 considers the total length of the source text and the number of words that should be kept and deleted together. Considering such additional length information can enable InstructCMP to recognize the length constraint more effectively. 
The number of words that should be kept is automatically calculated from the target desired length. 

Constraint 3-1 applies the \quotes{length priming} only to the source text based on its length, whereas Constraint 3-2 applies it solely to the target text based on the number of words that should be kept and deleted together.


\subsection{Dataset Creation}
We consider four benchmark datasets. 
The Google dataset (\textbf{Google}) was automatically created by considering the syntactic dependency tree structure from news headlines~\cite{filippova-altun-2013-overcoming}. The training, validation, and test datasets consist of 200,000, 1,000, and 1,000 pairs, respectively.
For the test dataset used in the evaluation, the gold compression ratio is 0.45.
The Broadcast (\textbf{Broad}) and BNC (\textbf{BNC}) datasets~\cite{Clarke2008GlobalIF} comprise manually 
compressed sentences. Each of these datasets 
contains 1,370 and 1,629 evaluation pairs, respectively. The gold compression ratios of these datasets, which are 0.76 and 0.72 respectively, are longer than those of other evaluation datasets. 
DUC2004 (TASK1) (\textbf{DUC}) comprises 500 pairs with a gold compression ratio of 0.39. Unlike other evaluation datasets, this dataset includes abstract summaries as its ground truth.

We created new datasets by transforming traditional sentence compression datasets into an instruction 
format. 
For length constraint instructions, we inject lengths of ground-truth summaries. 

\subsection{Instruction-based Fine-tuning}
To improve performances by leveraging LLM's generalizability~\cite{wang-etal-2022-super,weifinetuned,chung2022scaling}, we also created a training dataset for instruction-based fine-tuning by sampling 5\% of the training dataset from \textbf{Google}. 
Through this fine-tuning, we aim to enhance a model for better learning and improving abilities to handle length constraints in compressing sentences without any model modifications. 

\section{Experiments}
\subsection{Experimental Settings}

\noindent \textbf{Evaluation Metrics.} 
F$_1$ scores of ROUGE-1 (R-1), -2 (R-2), and -L (R-L), the F$_1$-score for kept tokens (F$_1$), and the BERT score (BS)~\cite{bert-score} were used to evaluate compression quality. 
The ROUGE scores were calculated using the implementation provided by Google Research.\footnote{\url{https://github.com/google-research/google-research/tree/master/rouge}}

To evaluate performances related to a length constraint, we calculated \textit{$\Delta CR$}, the difference between 
the model-generated compression ratio and the gold compression ratio.
\textit{$\Delta CR$} evaluates how close the compression ratio of model-generated outputs is to the gold compressed summary~\cite{kamigaito-etal-2018-higher,Kamigaito_Okumura_2020}. 
Because InstructCMP can produce novel words, we counted the number of the novel words in the model-generated compressed summaries. Thus, \textit{novel} represents the ratio of novel words that do not appear in the source text.

\begin{table*}[t!]
\renewcommand{\arraystretch}{1.2}
\begin{adjustbox}{width=1.9\columnwidth,center}

\centering

\small
\begin{tabular}{ccccccccccc}
\toprule
\textbf{Dataset} & \textbf{Setting} & \textbf{Instruction} & \textbf{Prompting} & \textbf{R-1} & \textbf{R-2} & \textbf{R-L} & \textbf{F$_1$} & \textbf{BS} & \textbf{$\Delta$ \textit{CR}} & \textbf{\textit{novel}} \\
\midrule
\multirow{8.5}{*}{Google} &     \multirow{4}{*}{Zero-shot}        & \#1  &    \xmark & \underline{65.88} & 55.48 & 65.42 & \underline{0.66} & 0.66 & \underline{+30.22} &  0.28\\  
     &  & \#2 &  Chain-of-Thought & 65.74 & \underline{56.12} & \underline{65.56} & 0.66 & 0.66 & +32.46 & 0.11\\
  &  & \#2 &  Tree-of-Thought & 65.56 & 55.34 & 65.19 & 0.66 & 0.66 & +30.99 & 0.17\\
      & & \#3 & Priming & \textbf{74.59}$^\dagger$ & \textbf{62.45}$^\dagger$ & \textbf{73.69}$^\dagger$ & \textbf{0.74}$^\dagger$ & \textbf{0.73} & +10.13$^\dagger$ &  0.57\\
\cmidrule{2-11}
&     \multirow{4}{*}{QLoRA fine-tuning}                   &  \#1      & \xmark & 82.85 & 75.15 & 82.58 & 0.84 & 0.82 & -1.28 &  0.17\\

        & & \#2 &  Chain-of-Thought & \underline{84.88} & \underline{77.20} & \underline{84.56} & \underline{0.86} & 0.83 & \underline{-0.90} & 0.18 \\
  & & \#2 &  Tree-of-Thought & 84.69 & 76.89 & 84.26 & 0.85 & 0.83 & -1.90 & 0.17\\
         &    &  \#3  & Priming & \textbf{86.88}$^\dagger$ & \textbf{79.55}$^\dagger$ & \textbf{86.26}$^\dagger$ & \textbf{0.87}$^\dagger$ & \textbf{0.84} &  -0.16$^\dagger$ &  0.17\\
\midrule

                     &\multirow{4}{*}{Zero-shot} &   \#1 &      \xmark & \underline{79.30} & 65.54 & \underline{78.27} & \underline{0.79} & \textbf{0.76} & +4.21 &  0.32\\

         & & \#2 &  Chain-of-Thought & 78.94 & \underline{65.76} & 78.21 & 0.79 & 0.75 & \underline{+3.99} & 0.19\\
  & & \#2 &  Tree-of-Thought  & 78.02 & 63.90 & 77.32 & 0.78 & 0.74 & +4.17 & 0.33\\
   \multirow{2.5}{*}{Broad}  & & \#3 & Priming & \textbf{80.27}$^\dagger$ & \textbf{66.62}$^\dagger$ & \textbf{79.30}$^\dagger$ & \textbf{0.80}$^\dagger$ & \textbf{0.76} &  -0.01$^\dagger$ &  0.33\\
\cmidrule{2-11}
  &  \multirow{4}{*}{QLoRA fine-tuning}          &       \#1      &  \xmark & 70.14 & 58.15 & 69.70 & 0.68 & 0.68 & -15.88 &  0.34 \\

   & & \#2 &  Chain-of-Thought & \underline{78.24} & \underline{65.61} & \underline{77.78} & \underline{0.77} & 0.72 & \underline{-3.96} & 0.36 \\
  & & \#2 &  Tree-of-Thought & 77.68 & 64.94 & 77.06 & 0.76 & 0.71 & -7.46 & 0.32 \\
  &    &  \#3  &  Priming & \textbf{82.63}$^\dagger$ & \textbf{69.76}$^\dagger$ & \textbf{81.16}$^\dagger$ & \textbf{0.81}$^\dagger$ & \textbf{0.75} & -1.38$^\dagger$ &  0.35\\
\midrule

          &   \multirow{4}{*}{Zero-shot}          &  \#1 &   \xmark & \underline{74.81} & 61.21 & 73.64 & 0.75 & \textbf{0.70} &  +10.38 &  0.37\\  

     &  & \#2 &  Chain-of-Thought & 74.46 & 61.03 & \underline{73.66} & \underline{0.75} & 0.69 & \underline{+3.57} & 0.11\\
  & & \#2 &  Tree-of-Thought  & 73.81 & 60.11 & 72.82 & 0.74 & 0.68 & +7.01 & 0.26\\
   \multirow{2.5}{*}{BNC}      & &  \#3 &  Priming & \textbf{75.78}$^\dagger$ & \textbf{61.76} & \textbf{74.52}$^\dagger$ & \textbf{0.76}$^\dagger$ & \textbf{0.70} &  +0.16$^\dagger$ &  0.25\\
\cmidrule{2-11}
        &  \multirow{4}{*}{QLoRA fine-tuning}            &   \#1       &  \xmark & 61.28 & 49.61 & 60.51 & 0.60 & 0.59 &  -24.21 &  0.27\\

         & & \#2 &  Chain-of-Thought & \underline{75.58} & \underline{62.55} & \underline{74.76} & \underline{0.74} & 0.68 &  -4.35 & 0.27\\
         & & \#2 &  Tree-of-Thought & 73.37 & 60.22 & 72.30 & 0.72 & 0.66 & -10.81 & 0.25 \\
                 &  &   \#3  &  Priming & \textbf{77.54}$^\dagger$ & \textbf{64.38}$^\dagger$ & \textbf{76.00}$^\dagger$ & \textbf{0.76}$^\dagger$ & \textbf{0.70} &  -4.13 &  0.26\\

\midrule

 & \multirow{4}{*}{Zero-shot} & \#1 &     \xmark & \underline{27.09} & \underline{8.72} & \underline{22.65} & \underline{0.23} & 0.33 & \underline{+37.97} &  0.25\\
        & & \#2 &   Chain-of-Thought & 26.28 & 8.35 & 21.86 & 0.23 & 0.32 & +40.53 & 0.10\\
  & & \#2 &          Tree-of-Thought & 26.13 & 8.20 & 21.75 & 0.23 & 0.32 & +40.31 & 0.19\\
 \multirow{2.5}{*}{DUC} & & \#3 & Priming & \textbf{28.19}$^\dagger$ & \textbf{9.66}$^\dagger$ & \textbf{24.56}$^\dagger$ & \textbf{0.24}$^\dagger$ & \textbf{0.34} & +15.08$^\dagger$ &  0.81\\
\cmidrule{2-11}
& \multirow{4}{*}{QLoRA fine-tuning} & \#1 &  \xmark & \textbf{27.31} & \textbf{9.21} & \textbf{24.34} & \textbf{0.24} & \textbf{0.35} & +0.28 &  0.18\\  
&   & \#2 &             Chain-of-Thought & 26.29 & 8.62 & 23.40 & 0.23 & 0.34 & -3.10 & 0.19 \\
&   & \#2 &             Tree-of-Thought  & 26.28 & 8.38 & 23.58 & 0.23 & 0.34 & -2.29 & 0.20 \\
&   & \#3 &                      Priming & 26.83 & 8.57 & 23.96 & 0.23 & 0.33 & +0.78 &  0.21\\
\bottomrule
\end{tabular}

\end{adjustbox}

\caption{Experimental results of InstructCMP using Llama2-13B-chat on \texttt{Google}, \texttt{Broad}, \texttt{BNC}, and \texttt{DUC}. Checkmark indicates not applying 
a length constraint. $\dagger$ indicates the improvement is significant (\textit{p}<0.05) compared with the underlined (generally, the best baseline score) on each dataset.}
\label{tab:unsupervised}
\end{table*}
 
\noindent \textbf{Implementation Details.}
We employed the instruction-based open-source Llama2-13B-chat model~\cite{touvron2023llama}\footnote{\url{https://huggingface.co/meta-llama/Llama-2-13b-chat-hf}} as our backbone model. We tested various instructions on the validation dataset from \texttt{Google} and made selections based on their performance. To explore various  parameter numbers, we experimented with 4-bit and 8-bit quantizations, as well as without quantization~\cite{Jacob_2018_CVPR} using PyTorch.\footnote{\url{https://github.com/pytorch/pytorch}} 
We also evaluated the performance across various model sizes, including 7B and 70B, and compared various model types, specifically the encoder-decoder based models of FLAN-T5-XXL (11B)~\cite{chung2022scaling}\footnote{\url{https://huggingface.co/google/flan-t5-xxl}} and FLAN-UL2 (20B)~\cite{tay2023ul2}.\footnote{\url{https://huggingface.co/google/flan-ul2}}

For instruction-based fine-tuning, we considered QLoRA, which can preserve the full 16-bit fine-tuning performance~\cite{dettmers2023qlora}. QLoRA is an extended version of Low-Rank Adapters (LoRA)~\cite{hu2022lora}, an improved Parameter-Efficient Fine-Tuning (PEFT)~\cite{peft} method for LLMs. This method combines low-rank and trainable matrices with the frozen weights in each layer of Transformer, building upon the foundational approach of LoRA. We incorporated low-rank matrices into the query and value weights using a LoRA attention dimension of 8. During training, we used 8-bit quantization for QLoRA, and during inference, we employed 4-bit quantization. 

\subsection{Main Results}
Table~\ref{tab:unsupervised} shows the performances of InstructCMP based on the Llama2-13B-chat model in a zero-shot setting, used directly without additional training, and in the QLoRA instruction-tuning setting, which involves fine-tuning of InstructCMP. Because prompting techniques for LLMs, such as few-shot~\cite{min-etal-2022-rethinking}, directional stimulus~\cite{li2023guiding}, and generated knowledge~\cite{liu-etal-2022-generated} methods, require external resources, we compared \quotes{length priming} to prompting techniques of chain-of-thought~\cite{COT} and tree-of-thought in a single prompt~\cite{yao2023tree,tree-of-thought-prompting} by adding them at the beginning of length constraint instructions (\#2 in Table~\ref{tab:instruction_format}).

\noindent \textbf{Performance in Instruction-based Zero-shot.}\footnote{Experiments considering various instructions on the validation dataset from \texttt{Google} are detailed in Appendix~\ref{sec:appendixA}.} 
Even in a zero-shot setting, InstructCMP without a length constraint (\#1 in Table~\ref{tab:instruction_format}) successfully compresses sentences, while it cannot necessarily meet the length.
In applying length constraints with \quotes{length priming}, consistently improved performances are observed in both ROUGE and $\Delta$\textit{CR}. 
In addition, our \quotes{length priming} significantly outperforms other prompting methods, chain-of-thought and tree-of-thought, in both length controllability and ROUGE metrics.

However, controlling the length of outputs for \texttt{Google} and \texttt{DUC} proved to be more challenging than \texttt{Broad} and \texttt{BNC}, specifically, in a zero-shot setting. We think this challenge arises from the nature of datasets, whose compression ratio is lower. Table~\ref{tab:length effect} shows the results based on a target compression ratio of 0.2 and a target word count of 5 words, respectively.
We observed that when the compression ratio is lower, the LLMs have difficulties maintaining both informativeness and length controllability.

\begin{table*}[t!]
\renewcommand{\arraystretch}{1.2}
\begin{adjustbox}{width=1.6\columnwidth,center}

\centering

\small

\begin{tabular}{ccccccccccc}
\toprule
\textbf{Data} & \textbf{Boundary} & \textbf{cnt} & \textbf{R-1} & \textbf{R-2} & \textbf{R-L} & \textbf{F$_1$} & \textbf{$\Delta$ \textit{CR}} & \textit{src len} & \textit{tgt len} & \textit{gen len}\\
\midrule
\multirow{10.5}{*}{Google} 
& 0.8$\sim$1.0 & 32  & 86.05 & 74.22 & 85.18 & 0.85 & -1.02 & - & - & -\\
& 0.6$\sim$0.8 & 180 & 81.09 & 69.64 & 79.96 & 0.80 &  8.24 & - & - & -\\
& 0.4$\sim$0.6 & 343 & 77.86 & 66.87 & 77.15 & 0.78 & 10.96 & - & - & -\\
& 0.2$\sim$0.4 & 403 & 70.15 & 56.94 & 69.15 & 0.69 & 11.05 & - & - & -\\
& 0.0$\sim$0.2 & 42  & 53.78 & 39.52 & 53.36 & 0.51 & 11.18 & - & - & -\\
\cmidrule{2-11}
& 20$\sim$   & 13  & 80.43 & 69.01 & 79.67 & 0.79 & - & 38.08 & 20.85 & 25.31 \\
& 15$\sim$20 & 127 & 78.22 & 66.98 & 76.86 & 0.77 & - & 29.46 & 16.31 & 18.58 \\
& 10$\sim$15 & 518 & 75.97 & 64.40 & 75.03 & 0.76 & - & 26.74 & 11.68 & 14.75 \\
& 5$\sim$10  & 338 & 71.11 & 57.75 & 70.45 & 0.70 & - & 25.90 &  7.69 & 10.16 \\
& 0$\sim$5   & 4   & 55.42 & 43.45 & 55.52 & 0.58 & - & 27.25 &  4.00 &  7.75 \\
\midrule
\multirow{8.5}{*}{DUC} 
& 0.8$\sim$1.0 & 8   & 11.23 & 3.43 & 10.26 & 0.15 & -9.44 & - & - & -\\
& 0.6$\sim$0.8 & 20  & 18.18 & 5.27 & 15.36 & 0.16 & 15.43 & - & - & -\\
& 0.4$\sim$0.6 & 118 & 30.51 & 10.48 & 26.12 & 0.27 & 14.60 & - & - & -\\
& 0.2$\sim$0.4 & 326 & 29.56 & 10.24 & 25.93 & 0.24 & 17.64 & - & - & -\\
& 0.0$\sim$0.2 & 18  & 20.61 &  5.91 & 17.88 & 0.18 & 11.86 & - & - & -\\
\cmidrule{2-11}
& 15$\sim$20 & 26  & 22.97 &  5.96 & 18.45 & 0.24 & - & 32.15 & 15.65 & 19.96 \\
& 10$\sim$15 & 363 & 29.95 & 10.23 & 26.07 & 0.26 & - & 33.55 & 11.62 & 17.06 \\
& 5$\sim$10  & 101 & 25.66 &  9.37 & 22.81 & 0.19 & - & 33.06 &  8.38 & 14.11 \\

\bottomrule
\end{tabular}
\end{adjustbox}
\caption{Effect of compression ratio and word count. \textit{cnt} indicates the number of instances in each boundary.}
\label{tab:length effect}
\end{table*}

\begin{table}[t!]
\begin{adjustbox}{width=0.9\columnwidth,center}
\centering
\small
\begin{tabular}{cccccccc}
\toprule
\textbf{Data}& \textbf{Setting} & \textbf{Output} & \textbf{Gram.} & \textbf{Faith.} & \textbf{Info.}\\
\midrule

\multirow{3}{*}{Google}& QLoRA& 13B     &\textbf{4.14}$^\dagger$ & 4.09 & \textbf{4.06}$^\dagger$ \\
                      & Zero-shot & 13B     & \underline{4.06} & 4.09 & \underline{4.00} \\ [0.5ex]
\cdashline{2-6}\noalign{\vskip 0.5ex}
                       & Gold & - & 4.03 & \textbf{4.11} & 4.05 \\
\midrule
\multirow{3.5}{*}{Broad} & \multirow{2}{*}{Zero-shot} & 13B & \textbf{3.92} & \textbf{3.88}  & 3.86 \\
                       &  & 70B     & 3.90  & 3.87   & \textbf{3.90}$^\dagger$ \\ [0.5ex]
\cdashline{2-6}\noalign{\vskip 0.5ex}                     & Gold & - & \textbf{3.92}  & \textbf{3.88}  & \underline{3.85} \\
\midrule
\multirow{3.5}{*}{BNC}   & \multirow{2}{*}{Zero-shot}& 13B     & \textbf{3.98} & 3.93 & 3.93 \\
                      &  & 70B     & 3.96 & 3.91 & \textbf{3.96} \\ [0.5ex]
\cdashline{2-6}\noalign{\vskip 0.5ex}                     & Gold & - & 3.96& \textbf{3.94} & 3.92 \\

\bottomrule
\end{tabular}
\end{adjustbox}
\caption{Human evaluation results. The notations are the same as those in Table~\ref{tab:unsupervised}.}
\label{tab:humaneval}
\end{table}

\noindent \textbf{Performance in Instruction-based Fine-tuning.}\footnote{Experiments considering 0.5\% and 1\% randomly sampled training datasets from \texttt{Google} are detailed in Appendix~\ref{sec:appendixB}.} 
Following instruction-based QLoRA fine-tuning, the created training dataset further improves performances of InstructCMP.
As shown in $\Delta$\textit{CR} for \texttt{Broad} and \texttt{BNC}, the model without the length constraint was trained to compress sentences more closely aligned with the gold compression ratio of \texttt{Google}.
However, the performance degradation was observed  on \texttt{DUC} when fine-tuning was applied using \texttt{Google}, due to the different natures of their abstractive and extractive ground-truth summaries. 
  
\noindent \textbf{Length Priming.}
The ablation results for \quotes{length priming} in instructions 
are presented in Table~\ref{tab:lengthpriming}.
We first compare performances of \quotes{length priming} in an unsupervised zero-shot setting.
It significantly improved performances on all datasets in terms of \textit{$\Delta CR$} compared to w/o priming. 
Even in a supervised instruction-based fine-tuning, \quotes{length priming} largely improved performances in both ROUGE metrics and length controllability. The exception is on \texttt{DUC} because of its nature of the abstractive gold summary.

\begin{figure}[t]
\centering
\includegraphics[width=7.5cm]{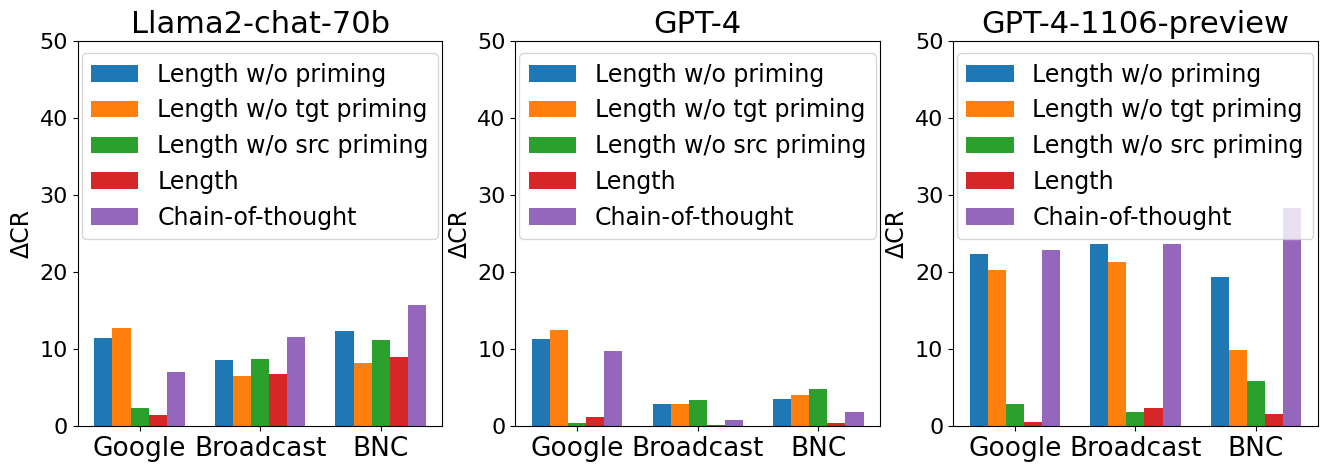}
\caption{Absolute $\Delta$\textit{CR} for \quotes{length priming} types.}
\label{fig:gpt4}
\end{figure}

We also compare the effectiveness of \quotes{length priming,} using larger models, such as Llama-2-70B-chat-hf, ChatGPT (GPT-4), and ChatGPT (GPT-4-1106-preview). 
Figure~\ref{fig:gpt4} shows the results. We confirm that \quotes{length priming} is essential for length constraints, even in the most recent and powerful LLMs.\footnote{When we additionally tested the chain-of-thought and tree-of-thought prompting methods on these larger models, their length controllability was similar to each other, which is similar to the results in Table~\ref{tab:unsupervised}.}

\begin{table}[t!]
\begin{adjustbox}{width=1\columnwidth,center}
\centering
\small

\begin{tabular}{ccccccccc}

\toprule
\textbf{Data} & \textbf{Method} & \textbf{Instruction} & \textbf{R-1} & \textbf{R-2} & \textbf{R-L} & \textbf{F$_1$} & \textbf{BS} & \textbf{$\Delta$ \textit{CR}} \\
\midrule

      &             &       \#2 & \underline{63.73} & \underline{54.04} & \underline{63.54} & \underline{0.64} & 0.64 &  \underline{+38.44}\\
 &      \multirow{2}{*}{Zero-shot}                       & \#3  & \textbf{74.59}$^\dagger$ & \textbf{62.45}$^\dagger$ & \textbf{73.69}$^\dagger$ & \textbf{0.74}$^\dagger$ & \textbf{0.73} & +10.13$^\dagger$\\

       &   & \#3-1 & 67.32 & 57.61 & 67.01 & 0.68 & 0.67 &  +30.63\\ 
\multirow{2.5}{*}{Google} &  & \#3-2 & 73.72 & 60.66 & 72.94 & 0.72 & 0.72 & +9.58\\
\cmidrule{2-9}
       &         &       \#2 & \underline{84.99} & \underline{77.43} & \underline{84.69} & \underline{0.86} & 0.83 &  \underline{+1.45}\\
       &    QLoRA           & \#3 & \textbf{86.88}$^\dagger$ & 79.55$^\dagger$ & 86.26$^\dagger$ & \textbf{0.87}$^\dagger$ & \textbf{0.84} &  -0.16$^\dagger$\\
       & fine-tuning & \#3-1 & 85.20 & 77.46 & 84.72 & 0.86 & 0.83 &  +0.76\\ 
       &        & \#3-2 & 86.80 & \textbf{79.58} & \textbf{86.29} & \textbf{0.87} & \textbf{0.84} &  +0.12\\

\midrule
         &             &       \#2 & 81.08 & 67.79 & \textbf{80.55} & \textbf{0.81} & \textbf{0.77} &  \underline{+8.78}\\
     &    \multirow{2}{*}{Zero-shot}          & \#3  & 80.27 & 66.62 & 79.30 & 0.80 & 0.76 &  -0.01$^\dagger$\\
                 &  & \#3-1 & \textbf{81.13} & \textbf{68.14} & \textbf{80.55} & \textbf{0.81} & \textbf{0.77} & +6.91\\ 
      \multirow{2.5}{*}{Broad}   &   & \#3-2 & 78.64 & 64.58 & 77.63 & 0.78 & 0.74 & -1.42\\

\cmidrule{2-9}
       &  & \#2       & \underline{80.34} & \underline{67.77} & \underline{79.81} & \underline{0.78} & 0.75 & -1.02\\
       &   QLoRA           & \#3  & 82.63$^\dagger$ & 69.76$^\dagger$ & 81.16$^\dagger$ & \textbf{0.81}$^\dagger$ & 0.75 & -1.38\\
              & fine-tuning  & \#3-1 & 82.80 & \textbf{70.39} & \textbf{82.05} & \textbf{0.81} & \textbf{0.77} &  +0.90\\ 
       &       & \#3-2 & \textbf{82.66} & 69.81 & 81.16 & \textbf{0.81} & 0.75 & -1.08\\

\midrule
    &             & \#2       & \textbf{77.36} & \textbf{63.64} & \textbf{76.59} & \textbf{0.78} & \textbf{0.72} & \underline{+10.46}\\

     &      \multirow{2}{*}{Zero-shot}       & \#3  & 75.78 & 61.76 & 74.52 & 0.76 & 0.70 & +0.16$^\dagger$\\
    &  & \#3-1 & 77.24 & 63.52 & 76.50 & 0.77 & \textbf{0.72} & +8.53\\ 
 
\multirow{2.5}{*}{BNC}    &    & \#3-2 & 73.16 & 59.17 & 71.82 & 0.73 & 0.68 & -4.05\\

\cmidrule{2-9}
       &  & \#2       & \underline{73.74} & \underline{61.52} & \underline{72.92} & \underline{0.72} & 0.68 &  \underline{-5.50}\\
       
       &   QLoRA          & \#3  & 77.54$^\dagger$ & 64.38$^\dagger$ & 76.00$^\dagger$ & 0.76$^\dagger$ & \textbf{0.70} &  -4.13$^\dagger$\\
       
      & fine-tuning  & \#3-1 & \textbf{77.62} & \textbf{64.58} & \textbf{76.45} & \textbf{0.77} & \textbf{0.70} &  -1.49\\

       &        & \#3-2 & 77.40 & 64.20 & 75.81 & 0.76 & 0.68 &  -4.03\\

\midrule
&       & \#2       & \underline{26.23} & \underline{8.38} & \underline{21.70} & \underline{0.23} & 0.31 & \underline{+46.37}\\
  &     \multirow{2}{*}{Zero-shot}        & \#3  & 28.19 $^\dagger$ & 9.66$^\dagger$ & 24.56$^\dagger$ & \textbf{0.24}$^\dagger$ & \textbf{0.34} & +15.08$^\dagger$\\
                         &  & \#3-1 & 26.53 & 8.59 & 22.33 & 0.23 & 0.32 & +41.51\\ 
            
\multirow{2.5}{*}{DUC}& & \#3-2 & \textbf{28.41} & \textbf{9.85} & \textbf{24.66} & \textbf{0.24} & \textbf{0.34} & +16.45\\

\cmidrule{2-9}
       &             & \#2       & \textbf{27.20} & \textbf{8.98} & \textbf{24.27} & \textbf{0.24} & \textbf{0.35} & +0.47\\
       &  QLoRA           & \#3   & 26.83 & 8.57 & 23.96 & 0.23 & 0.33 & +0.78\\
       & fine-tuning & \#3-1 & 26.25 & 8.27 & 23.49 & 0.23 & 0.34 & -1.22\\
       &        & \#3-2 & 26.46 & 8.31 & 23.62 & 0.23 & 0.33 & +1.32\\

\bottomrule
\end{tabular}

\end{adjustbox}

\caption{Ablation study for \quotes{length priming.} The notations are the same as those in Table~\ref{tab:unsupervised}. }
\label{tab:lengthpriming}
\end{table}

\section{Analysis}
\subsection{Parameter Sizes}
\begin{figure}[t]
\centering
\includegraphics[width=7cm]{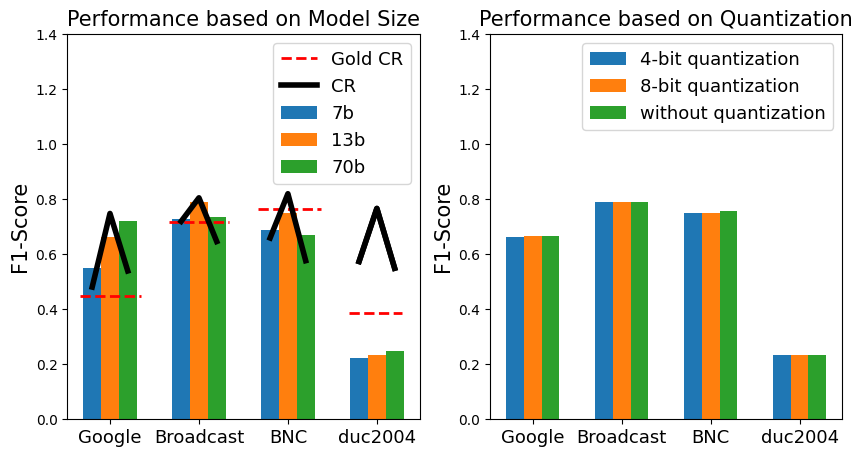}
\caption{Performances for different model sizes and quantizations.}
\label{fig:size_quant}
\end{figure}

The left graph of Figure~\ref{fig:size_quant} shows the F$_1$ score for kept tokens and the model-generated compression ratio (\textit{CR}), compared to the gold compression ratio, based on zero-shot InstructCMP without a length constraint on the Llama2-chat model with 7B, 13B, and 70B parameters.
On \texttt{Google} and \texttt{DUC}, the F$_1$ scores increased with enlarging the model size, achieving compression closer to the gold compression ratio. However, on \texttt{Broadcast} and \texttt{BNC}, which have high gold compression ratios, InstructCMP with the 70B model compresses sentences more concisely, resulting in a compression ratio that significantly deviates from the gold compression ratio, consequently decreasing F$_1$ scores compared to the 13B model. 

To further investigate this, we conducted human evaluations. We sampled 100 instances each from \texttt{Google}, \texttt{Broad}, and \texttt{BNC}. By using Amazon Mechanical Turk, we assigned in total 120 evaluators who obtained both US high school and US bachelor's degrees for grading the results with scores from 1 to 5 (5 is the best) in terms of grammatical correctness (Gram), factual consistency (Faith), and a balance of redundancy and informativeness (Info). Table~\ref{tab:humaneval} shows the results. Because of the automatically constructed nature of \texttt{Google}, QLora and zero-shot settings can yield higher grammaticality scores than the gold summary. 
These results also indicate gold summaries of \texttt{Broad} and \texttt{BNC} are actually redundant~\cite{ghalandari-etal-2022-efficient}, and our instruction-based approach can generate faithful, informative, and grammatical summaries. 

The right graph of Figure~\ref{fig:size_quant} shows the results of zero-shot InstrcutCMP without a length constraint on Llama2-13B-chat. Interestingly, there are no significant differences in performance among the 4-bit, 8-bit, and nonquantized versions.

\subsection{Model Types}
It is also of interest to draw comparisons with other instruction-based models, such as FLAN-T5-XXL and FLAN-UL2, both of which employ the encoder-decoder architecture. However, they did not effectively compress sentences using instruction templates in Table~\ref{tab:instruction_format}. We think this is due to the nature of their pre-training, which causes potential gaps between the pre-training steps and the instruction templates for extractive summarization settings~\cite{kwon-etal-2023-abstractive}. Thus, we used slightly modified instruction templates.\footnote{Experimental results using instruction templates in Table~\ref{tab:instruction_format} and modified instruction templates are in Appendix~\ref{sec:appendixC}.} 
Table~\ref{tab:flan} shows the results. Our \quotes{length priming} can improve length controllability by keeping ROUGE metrics compared to w/o priming.


\subsection{Case Study}
Table~\ref{tab:case} shows the outputs of zero-shot InstructCMP based on the Llama-13B-chat model. 
The first example shows the controllability of the length constraint instruction. 
Even when instructed to delete zero words, InstructCMP follows the instruction correctly. 
The second example shows the flawless grammatical capabilities of LLMs~\cite{mitrovi2023chatgpt}. When deleting a single word can cause a grammatical error, InstructCMP can correct the 
error by paraphrasing, represented as \textit{novel} in Table~\ref{tab:unsupervised}.
The third example shows the output of InstructCMP in response to the length constraint. \quotes{Length priming} assists InstructCMP to compress a source text to meet a desired length, performing better than the length constraint without priming. 

\begin{table}[t!]
\begin{adjustbox}{width=1\columnwidth,center}
\renewcommand{\arraystretch}{1.2}
\centering
\small

\begin{tabular}{cccccccc}

\toprule
\textbf{Data} & \textbf{Model} & \textbf{Instruction} & \textbf{R-1} & \textbf{R-2} & \textbf{R-L} & \textbf{F$_1$} & \textbf{$\Delta$ \textit{CR}}\\
\midrule

& \multirow{3}{*}{T5-XXL} & \#1 & 60.06 & 50.52 & 59.81 & 0.60 & +47.84\\
&                         & \#2 & \underline{62.41} & 51.18 & \underline{61.90} & \underline{0.61} & \underline{+35.72}\\
\multirow{2.5}{*}{Google} & & \#3 & \textbf{66.22}$^\dagger$ & \textbf{51.68} & \textbf{65.43}$^\dagger$ & \textbf{0.62}$^\dagger$ & +19.51$^\dagger$\\
\cmidrule{2-8}
& \multirow{3}{*}{UL2} & \#1 & 63.53 & \underline{45.79} & 62.35 & 0.57 & +11.92\\
&                      & \#2 & \underline{64.72} & 44.38 & \underline{63.87} & \underline{0.57} & +1.11\\
       &               & \#3 & \textbf{66.06}$^\dagger$ & \textbf{47.24}$^\dagger$ & \textbf{65.39}$^\dagger$ & \textbf{0.59}$^\dagger$ & +6.34\\
\midrule

& \multirow{3}{*}{T5-XXL} & \#1 & \textbf{82.45} & \textbf{69.33} & \textbf{81.93} & \textbf{0.81} & +12.72\\
&                         & \#2 & 74.42 & 59.57 & 72.96 & 0.72 & +2.18\\
\multirow{2.5}{*}{Broad} &  & \#3 & 77.68 & 63.47 & 76.58 & 0.76 & +4.78\\
\cmidrule{2-8}
& \multirow{3}{*}{UL2} & \#1 & 73.82 & \underline{56.45} & \underline{70.90} & \underline{0.70} & \underline{-7.77}\\
&                      & \#2 & 68.79 & 52.27 & 66.70 & 0.66 & -9.84\\
&                      & \#3 & \textbf{74.31} & \textbf{59.12}$^\dagger$ & \textbf{72.79}$^\dagger$ & \textbf{0.71}$^\dagger$ & -4.04$^\dagger$\\
\midrule
&  \multirow{3}{*}{T5-XXL} & \#1 & \textbf{75.35} & \textbf{61.44} & \textbf{74.33} & \textbf{0.74} & +11.30\\
&                          & \#2 & 63.99 & 48.06 & 61.90 & 0.61 & \underline{-5.48}\\
\multirow{2.5}{*}{BNC} &     & \#3 & 65.43 & 49.78 & 63.55 & 0.62 & -3.43$^\dagger$\\
\cmidrule{2-8}
&\multirow{3}{*}{UL2} & \#1 & \textbf{67.42} & \textbf{49.90} & \textbf{63.88} & \textbf{0.62} & \underline{-10.64}\\
         &            & \#2 & 60.40 & 43.54 & 57.45 & 0.56 & -13.62\\
                   &  & \#3 & 64.88 & 49.03 & 62.81 & 0.61 & -8.17$^\dagger$\\
\bottomrule
\end{tabular}
\end{adjustbox}
\caption{Experimental results from zero-shot instruction-based FLAN models using encoder-decoder architectures. The notations are the same as those in Table~\ref{tab:unsupervised}.}
\label{tab:flan}
\end{table}

\begin{table}[t!]
\begin{adjustbox}{width=0.95\columnwidth,center}
\begin{tabular}{l}\toprule
\textbf{Source.} Eni has won a license for exploration block SM-857  offshore \\ Brazil.\\
\textbf{Instruction.} Sentence that consists of 11 words:\textbackslash n\{source\}\textbackslash nThe \\sentence that consists  of 11 words without the less important 0 words \\would be\textbackslash n:\\
\textbf{InstructCMP.} Eni has won a license for exploration block SM-857 \\offshore Brazil.\\
\midrule
\textbf{Source.} Rick Riordan has revealed the cover for his latest crossover \\short story, 
\quotes{Staff of Serapis}, which features Annabeth Chase and \\Sadie Kane.\\
\textbf{InstructCMP.} Rick Riordan has revealed the cover for his latest \\crossover short story, 
featuring Annabeth Chase and Sadie Kane.\\
\midrule
\textbf{Source.} Chinese shares closed lower Wednesday dragged down by \\the bio-pharmaceutical sector and small enterprises with growth \\potential.\\
\textbf{Length const. w/o priming.} Chinese shares closed lower Wednesday \\dragged down by the bio-pharmaceutical sector.\\
\textbf{Length const.} Chinese shares closed lower Wednesday.\\
\textbf{Gold}: Chinese shares closed lower Wednesday.\\
\bottomrule
\end{tabular}

\end{adjustbox}
\caption{Outputs of InstructCMP on \texttt{Google}. }
\label{tab:case}

\end{table}

\begin{table}[t!]
\begin{adjustbox}{width=1\columnwidth,center}
\centering
\small
\begin{tabular}{lccccccc}
\toprule
\textbf{Data} & \textbf{Model}  & \textbf{R-1} & \textbf{R-2} & \textbf{R-L} & \textbf{F$_1$} & \textbf{BS} & \textbf{\textit{len}}\\\midrule
\multicolumn{8}{c}{\textbf{Unsupervised}}\\\midrule
\multirow{3.5}{*}{Google}  &      SCRL$^\ast$           & 70.22 & 53.03  & 69.84  & 0.71 & &  10.8\\[0.5ex]
\cdashline{2-8}\noalign{\vskip 1.0ex}   
 & SCRL & \underline{70.53} & \underline{53.30} & \underline{70.07} & \underline{0.71 }& 0.65 & \underline{10.3}\\
& InstructCMP            & \textbf{74.92}$^\dagger$ & \textbf{62.53}$^\dagger$ & \textbf{73.83}$^\dagger$ & \textbf{0.75}$^\dagger$ & \textbf{0.75}  & 10.8$^\dagger$\\
\midrule
\multirow{2}{*}{Broad}       & SCRL               & \textbf{83.04} & \textbf{66.64} & \textbf{82.64} & \textbf{0.82} & \textbf{0.74} &  \underline{81\%} \\
  & InstructCMP        & 77.93 & 63.33 & 76.85 & 0.78 & \textbf{0.74} &  77\%$^\dagger$ \\
\midrule
\multirow{2}{*}{BNC}     &SCRL                & \textbf{79.55} & \textbf{62.24} & \textbf{78.69} & \textbf{0.79} & 0.69 &  \underline{79\%}\\
  & InstructCMP        & 75.11 & 60.56 & 74.03 & 0.75 & \textbf{0.70} & 74\%$^\dagger$\\
\midrule
\multirow{2}{*}{DUC}    & SCRL                  & \underline{26.78} & \underline{8.14} & \underline{23.30} & \underline{0.22} & 0.25 & \underline{10.0}\\
    & InstructCMP    & \textbf{28.14}$^\dagger$ & \textbf{9.43}$^\dagger$ & \textbf{24.82}$^\dagger$ & \textbf{0.23}$^\dagger$ & \textbf{0.32} & 10.6$^\dagger$\\
\midrule
\multicolumn{8}{c}{\textbf{Supervised}}\\\midrule
\multirow{3.5}{*}{Google} & SLAHAN$^\ast$ &  &  &  & 0.86 &  & \\ [0.5ex]
\cdashline{2-8}\noalign{\vskip 1.0ex}
 & SLAHAN & \textbf{82.98} & \underline{74.35} & \textbf{82.75} & \underline{0.83} & 0.78 & 9.3\\
         & InstructCMP & 82.85 & \textbf{75.15}$^\dagger$ & 82.58 & \textbf{0.84}$^\dagger$ & \textbf{0.82} &  9.5\\
\bottomrule
\end{tabular}
\end{adjustbox}
\caption{
Comparison with traditional state-of-the-art baselines. 
 $\ast$ indicates the reported score in the original paper. \textit{len} indicates the generated summary length. The notations are the same as those in Table~\ref{tab:unsupervised}.}
\label{tab:sota}
\end{table}

\subsection{Comparison with the Baselines} 
We compare InstructCMP with traditional state-of-the-art (SOTA) baselines, specifically \textbf{SCRL},\footnote{\url{https://github.com/complementizer/rl-sentence-compression}} which employs reinforcement learning optimized in unsupervised settings, and \textbf{SLAHAN},\footnote{\url{https://github.com/kamigaito/SLAHAN}} which recursively tracks parent and child words and leverages BERT embeddings optimized in supervised settings, trained on  \texttt{Google}~\cite{Kamigaito_Okumura_2020}. 

Following 
SCRL, we set a desired length of 11 for \texttt{Google} and \texttt{DUC}. In line with the previous work, we truncated model-generated outputs to 75 characters and used ROUGE recall scores for \texttt{DUC}~\cite{schumann-etal-2020-discrete,ghalandari-etal-2022-efficient}. For \texttt{Broadcast} and \texttt{BNC}, the desired length was set to 75\% of the length of the source sentence. Table~\ref{tab:sota} shows the results. 
Because zero-shot InstructCMP faces challenges in compressing sentences with length constraints when the gold compression ratio is low, we increased the model capability by using Llama2-70B-chat for \texttt{Google} and \texttt{DUC} instead of Llama2-13B-chat. We observed comparable performances of InstructCMP to SCRL.

We also compare InstructCMP, based on Llama-13B-chat, with SLAHAN. Following the previous work, we fine-tuned InstructCMP without a length constraint and achieved significant improvement, even after using 5\% of the training dataset.

\subsection{Increasing Training Dataset Size}
We provide additional experimental results using larger datasets for QLoRA fine-tuning with 10\% and 15\% google training datasets. Table~\ref{tab:LoRA_15} shows the results. Three different benchmark results on \texttt{Google}, \texttt{Broad}, and \texttt{BNC} support that length priming is necessary, except for DUC due to its abstract summary nature, and indicate the generalization of the length priming instruction.

\begin{table}[t!]
\begin{adjustbox}{width=0.9\columnwidth,center}
\centering
\small
\begin{tabular}{ccccccc}
\toprule
\textbf{Data}& \textbf{Size} & \textbf{R-1} & \textbf{R-2} & \textbf{R-L} & \textbf{F$_1$} & \textbf{$\Delta$\textit{CR}}\\
\midrule
Google & \multirow{4}{*}{10\%} & 87.45 & 80.47 & 87.00 & 0.88 & 0.69 \\
Broad  &                       & 79.21 & 66.31 & 77.51 & 0.78 & -1.44 \\
BNC    &                       & 83.38 & 70.29 & 81.84 & 0.81 & 0.42 \\
DUC    &                       & 27.02 &  8.34 & 23.85 & 0.23 & 2.09 \\
\midrule
Google & \multirow{4}{*}{15\%} & 89.01 & 82.24 & 88.56 & 0.89 & 0.39 \\
Broad  &                       & 79.72 & 66.47 & 78.27 & 0.79 & 0.02 \\
BNC    &                       & 82.92 & 69.65 & 81.90 & 0.82 & 0.57 \\
DUC    &                       & 26.30 &  7.92 & 23.53 & 0.23 & 2.14 \\
\bottomrule
\end{tabular}
\end{adjustbox}
\\
\caption{LoRA fine tuned model: training dataset size 10\% and 15\% from randomly sampled from Google dataset with \#3 instruction based on the 13B model}
\label{tab:LoRA_15}
\end{table}

\section{Related Work}
\noindent \textbf{Sentence Compression.} 
Early studies on sentence compression in both supervised and unsupervised learning frameworks have used linguistic constraints, such as tree trimming methods~\cite{jing-2000-sentence,knight,hori,clarke-lapata-2006-models,berg-kirkpatrick-etal-2011-jointly,filippova-altun-2013-overcoming}. To avoid potential parsing errors in the tree trimming, LSTM-based models have been introduced for deletion-based compression~\cite{filippova-etal-2015-sentence} by jointly considering eye-tracking data~\cite{klerke-etal-2016-improving} and incorporating a score function of an ILP-based tree trimming method~\cite{wang-etal-2017-syntax}. \newcite{zhao-etal-2018-language} explored reinforcement learning for a syntax-based language model, that does not use explicit parsed trees. \newcite{kamigaito-etal-2018-higher,Kamigaito_Okumura_2020} proposed Seq2Seq approaches that jointly learn sentence compression and dependency trees within their attention networks inspired by supervised head attention \cite{kamigaito-etal-2017-supervised}, an extensible approach to document-level summarization \cite{ishigaki-etal-2019-discourse} similar to the case of graph neural networks \cite{xu-etal-2020-discourse,kwon-etal-2021-considering}. Alternatively, some recent work has utilized LLMs, such as BERT, for sentence compression to optimize fluency in unsupervised frameworks~\cite{zhou-rush-2019-simple,Deleter,schumann-etal-2020-discrete}. Because a high-quality compressed sentence can infer from the original sentence, encoder-decoder-based autoencoder approaches have been also explored~\cite{miao-blunsom-2016-language,fevry-phang-2018-unsupervised,malireddy-etal-2020-scar}. For better optimization, reinforcement learning has been  used~\cite{wangrl,ghalandari-etal-2022-efficient}.

\noindent \textbf{Length Control.} 
Despite the success of previous studies, practical summarization requires additional constraints such as a summary length for compressing sentences~\cite{liu-etal-2018-controlling,takase-okazaki-2019-positional,Li_Zhu_Zhang_Zong_He_2020,he-etal-2022-ctrlsum}. The approach for controlling the output to a desired length required modifying model parameters~\cite{kikuchi-etal-2016-controlling}, applying direct constraints~\cite{takase-okazaki-2019-positional,makino-etal-2019-global,kwon-etal-2023-abstractive}, or splitting the training dataset into specific length ranges~\cite{he-etal-2022-ctrlsum} due to the limited model abilities. Traditionally, sentence compression heavily relies on the model modifications for constraints such as lengths~\cite{schumann-etal-2020-discrete,ghalandari-etal-2022-efficient}. 

\noindent \textbf{Instruction-based LLMs.} 
LLMs can perform various tasks in a zero-shot setting, using instruction-formatted inputs~\cite{NEURIPS2020_1457c0d6,Radford2019LanguageMA}. The emergence of instruction-based LLMs, such as ChatGPT 
and GEMINI,\footnote{\url{https://gemini.google.com/}} has demonstrated a significant improvement in performance, particularly in their zero-shot problem-solving abilities~\cite{feng2023sentence,fang2023chatgpt}. Because performance varies greatly with various instructions, previous studies focused on finding better instructions~\cite{zhu2023multilingual,wang2023large,yao2023tree}. Various prompting methods have been investigated, such as few-shot, directional stimulus, generated knowledge, chain-of-thought, and tree-of thought~\cite{min-etal-2022-rethinking,li2023guiding,liu-etal-2022-generated,COT,yao2023tree}. These new types of LLMs mark the beginning of a new era in the field of natural language processing.

While the capabilities of these LLMs continue to grow with an increasing number of parameters, challenges are introduced for these models in training and testing steps to provide robust and generalized outputs~\cite{rae2022scaling,smith2022using,chowdhery2022palm,chung2022scaling,NEURIPS2020_1457c0d6,tay2023ul2}. To address this issue, PEFT methods such as LoRA have been introduced. These methods combine low-rank and trainable matrices with frozen weights in each layer of Transformer and even consider quantization~\cite{hu2022lora,dettmers2023qlora}.

As a related approach to priming, label embedding \cite{xiong-etal-2021-fusing,zhang-etal-2021-language} can also incorporate label-related information into the input to enhance generation, as mentioned by \citet{kwon-etal-2023-hierarchical}. However, in contrast to priming, label embedding cannot precisely control the generation itself and requires additional training.

To conduct the sentence compression task with instructions, we focus on priming that incorporates additional constraint-specific information to enhance performance, particularly for the length constraint, rather than just paraphrasing instructions to direct the task. 

\section{Conclusion}
We proposed InstructCMP to conduct sentence compression by incorporating length constraints without model modifications. For this new approach, we constructed new evaluation datasets by transforming traditional sentence compression datasets into an instruction format, while we also created new training datasets. Additionally, we introduced \quotes{length priming} into the instructions and demonstrated its effectiveness in zero-shot and instruction-based fine-tuning settings on four benchmark datasets. 
We also conducted an in-depth analysis, including the model size and type. 

\section*{Limitations}
Although our length priming successfully compresses sentences, it might be challenging to consider it in document summarization, which requires considering multiple sentences. Therefore, it remains a topic for future studies. In the future, we will consider sentence relationships for prompting to summarize documents.
Furthermore, there can be cases where keyword constraints are required for controllable summarization to take into account the content of summaries, which also remains a potential area for future investigation.

\section*{Acknowledgements}
We would like to gratefully acknowledge the anonymous reviewers for their helpful comments and feedbacks.

\bibliography{anthology,custom}
\bibliographystyle{acl_natbib}

\appendix

\section{Performance in Instruction Selection}
\label{sec:appendixA}
To determine task-specific instructions, we manually composed several candidates and evaluated their performances on the validation dataset from \texttt{Google}. Table~\ref{tab:googlevalid} shows the results. 
Based on their performances, we selected the 5th instruction as our base instruction for the setting without a length constraint. 

\begin{table*}[t!]
\begin{adjustbox}{width=2.0\columnwidth,center}
\centering
\small
\renewcommand{\arraystretch}{1.2}
\begin{tabular}{clccccccc}

\toprule
\# &\textbf{Instruction}& \textbf{R-1} & \textbf{R-2} & \textbf{R-L} & \textbf{F$_1$} & \textbf{$\Delta$ \textit{CR}} & \textbf{\textit{novel}} \\
\midrule

1&Sentence:\textbackslash n\{input\}\textbackslash nThe sentence without the non-essential words would be:\textbackslash n 
& 64.73 & 54.59 & 64.23 & 0.65 &  +35.76 &  0.36 \\

2&Sentence:\textbackslash n\{input\}\textbackslash nThe compressed version of the original sentence without
& 60.53 & 48.10 & 59.43 & 0.61 &  +38.09 &  1.12\\
&generating new words:\textbackslash n &&&&&\\

3&Sentence:\textbackslash n\{input\}\textbackslash nCompress the sentence by removing the non-essential words:\textbackslash n
& 62.37 & 50.95 & 61.61 & 0.62 &  +36.84 &  0.80\\

4 &Sentence:\textbackslash n\{input\}\textbackslash nDelte the non-essential words by keeping the original meaning:\textbackslash n
& 61.50 & 51.87 & 61.18 & 0.62 &  +43.23 &  0.35\\

5&Sentence:\textbackslash n\{input\}\textbackslash nThe sentence without the less important words would be:\textbackslash n
& \textbf{66.99} & \textbf{56.79} & \textbf{66.54} & \textbf{0.68} &  +30.21 &  0.25\\

6&Original Sentence:\textbackslash n\{input\}\textbackslash nMake a new sentence without the non-essential words.&&&&&\\ 
&New sentence would be:\textbackslash n
& 64.23 & 52.26 & 62.79 & 0.65 &  +31.62 &  0.92\\
7&Sentence:\textbackslash n\{input\}\textbackslash nThe sentence without the unnecessary words would be:\textbackslash n
& 63.85 & 53.33 & 63.29 & 0.64 &  +36.48 &  0.51\\








\bottomrule
\end{tabular}

\end{adjustbox}

\caption{Performances of different instructions using zero-shot InstructCMP based on the Llama2-chat-13B model on the validation dataset of \texttt{Google}.}
\label{tab:googlevalid}
\end{table*}

\section{Performance in Varying Training Dataset Size}
\label{sec:appendixB}

To investigate the impact of the training dataset size on performance, we also prepared 0.5\% and 1\% training datasets randomly sampled from the \texttt{Google} dataset. 
Table~\ref{tab:qloradata} shows the results of QLoRa fine-tuning for constraints in supervised settings. As observed, increasing the dataset size correlates with improved performance. 
Table~\ref{tab:qloradata2} shows the results of an ablation study on \quotes{length priming}. Similarly, our \quotes{length priming} proves to be essential for performance improvements even in small datasets.

\section{Performance of the FLAN Models and Modified Instruction Templates}
\label{sec:appendixC}
Table~\ref{tab:flan1} shows the results for the FLAN models using instruction templates in Table~\ref{tab:instruction_format}. They did not effectively compress sentences, as denoted by $\Delta$\textit{CR}. 

Table~\ref{tab:instruction_format2} shows the modified templates for the FLAN models.

\begin{table}[t!]
\begin{adjustbox}{width=1\columnwidth,center}
\centering
\small

\begin{tabular}{cccccccc}

\toprule
\textbf{Data} & \textbf{Size} & \textbf{Instruction} & \textbf{R-1} & \textbf{R-2} & \textbf{R-L} & \textbf{F$_1$} & \textbf{$\Delta$ \textit{CR}} \\
\midrule

\multirow{2}{*}{Google}  &  & \#1              & 80.50 & 72.22 & 80.22 & 0.81 & +1.49 \\
                        &  & \#3         & \textbf{83.56} & \textbf{75.33} & \textbf{82.97} & \textbf{0.84} & -0.21\\ 

\cmidrule{3-8}
\multirow{2}{*}{Broad}  &  & \#1              & 71.46 & 59.30 & 70.88 & 0.70 & -14.37 \\
  & \multirow{2}{*}{0.5\%} & \#3         & \textbf{80.62} & \textbf{68.34} & \textbf{79.31} & \textbf{0.79} &  -5.68\\ 
\cmidrule{3-8}
\multirow{2}{*}{BNC}     &  & \#1              & 64.28 & 52.43 & 63.42 & 0.63 & -19.59 \\
                         &  & \#3         & \textbf{73.49} & \textbf{60.88} & \textbf{72.07} & \textbf{0.72} & --10.89\\ 
\cmidrule{3-8}
\multirow{2}{*}{DUC}     &  & \#1      & \textbf{26.91} & \textbf{8.61} & \textbf{23.59} & \textbf{0.23} & +3.06 \\
                         &  & \#3 & 26.15 & 8.07 & 23.25 & \textbf{0.22} & +0.9\\

\midrule
\multirow{2}{*}{Google}  &  & \#1              & 81.68 & 73.55 & 81.40 & 0.83 & +2.05 \\
                         &  & \#3         & \textbf{85.45} & \textbf{77.55} & \textbf{84.83} & \textbf{0.86} & +0.46\\ 

\cmidrule{3-8}
\multirow{2}{*}{Broad}   &  & \#1              & 72.54 & 60.57 & 72.04 & 0.71 & -13.04 \\
& \multirow{2}{*}{1\%}    & \#3         & \textbf{82.25} & \textbf{69.63} & \textbf{80.62} & \textbf{0.80} &  -3.38\\
\cmidrule{3-8}
\multirow{2}{*}{BNC}     &  & \#1              & 64.63 & 52.80 & 63.74 & 0.64 & -18.93 \\
                        &  & \#3         & \textbf{76.49} & \textbf{63.46} & \textbf{74.73} & \textbf{0.75} &  -6.64\\ 
\cmidrule{3-8}
\multirow{2}{*}{DUC}      &  & \#1      & \textbf{27.69} & \textbf{8.95} & \textbf{24.24} & \textbf{0.24} & +3.77 \\
                          &  & \#3 & 26.63 & 8.57 & 23.93 & 0.23 & +1.73\\
\midrule
\multirow{2}{*}{Google} &  & \#1              & 82.85 & 75.15 & 82.58 & 0.84 & -1.28 \\
                        &  & \#3         & \textbf{86.88} & \textbf{79.55} & \textbf{86.26} & \textbf{0.88} & -0.16 \\ 

\cmidrule{3-8}
\multirow{2}{*}{Broad}  &  & \#1              & 70.14 & 58.15 & 69.70 & 0.68 & -15.88 \\
    & \multirow{2}{*}{5\%} & \#3         & \textbf{82.63} & \textbf{69.76} & \textbf{81.16} & \textbf{0.81} &  -1.38\\ 
\cmidrule{3-8}
\multirow{2}{*}{BNC}    &  & \#1              & 61.28 & 49.61 & 60.51 & 0.60 & -24.21 \\
                        &  & \#3         & \textbf{77.54} & \textbf{64.38} & \textbf{76.00} & \textbf{0.76} &  -4.13\\ 
\cmidrule{3-8}
\multirow{2}{*}{DUC}   &  & \#1      & \textbf{27.31} & \textbf{9.21} & \textbf{24.34} & \textbf{0.24} & +0.28 \\
                       &  & \#3 & 26.83 & 8.57 & 23.96 & 0.23 & +0.78\\

\bottomrule
\end{tabular}

\end{adjustbox}

\caption{Experimental results of InstructCMP using Llama2-13B-chat for different training dataset sizes on \texttt{Google}, \texttt{Broad}, \texttt{BNC}, and \texttt{DUC}.}
\label{tab:qloradata}
\end{table}

\begin{table}[t!]
\begin{adjustbox}{width=1\columnwidth,center}
\centering
\small

\begin{tabular}{cccccccc}

\toprule
\textbf{Data} & \textbf{Size} & \textbf{Instruction} & \textbf{R-1} & \textbf{R-2} & \textbf{R-L} & \textbf{F$_1$} & \textbf{$\Delta$ \textit{CR}} \\

\midrule
                        &  & \#2       & 80.35 & 72.20 & 80.08 & 0.81 & +1.79\\
\multirow{2}{*}{Google}                        &  & \#3  & \textbf{83.56} & \textbf{75.33} & \textbf{82.97} & \textbf{0.84} & -0.21\\
                        &  & \#3-1 & 81.38 & 72.68 & 81.00 & 0.82 & 0.00\\ 
 &  & \#3-2 & 83.23 & 74.67 & 82.73 & \textbf{0.84} & -0.64\\

\cmidrule{3-8}
                        &    & \#2       & 72.09 & 59.95 & 71.59 & 0.71 & -12.55\\
  \multirow{2}{*}{Broad}                      &  & \#3   & \textbf{80.62} & \textbf{68.34} & 79.31 & \textbf{0.79} & -5.68\\
                        &  & \#3-1 & 76.87 & 64.49 & 76.32 & 0.76 & -7.51\\ 
  & \multirow{2}{*}{0.5\%} & \#3-2 & 80.31 & 68.12 & \textbf{79.51} & \textbf{0.79} & -4.98\\

\cmidrule{3-8}
                        &  & \#2       & 64.89 & 52.96 & 63.95 & 0.64 & -10.53\\
 \multirow{2}{*}{BNC}                        &  & \#3   & \textbf{73.49} & \textbf{60.88} & \textbf{72.07} & \textbf{0.72} & -10.89\\
                        &  & \#3-1 & 70.81 & 58.33 & 69.81 & 0.70 & -12.57\\ 
   &  & \#3-2 & 72.76 & 60.32 & 71.48 & \textbf{0.72} & -0.85\\

\cmidrule{3-8}
                        &  & \#2       & \textbf{27.16} & \textbf{9.02} & \textbf{23.95} & \textbf{0.23} & +3.04\\
\multirow{2}{*}{DUC}                         &  & \#3   & 26.15 & 8.07 & 23.25 & 0.22 & +0.90\\
                        &  & \#3-1 & 25.51 & 8.16 & 22.68 & 0.22 & -0.43\\
   &  & \#3-2 & 26.33 & 8.29 & 23.65 & \textbf{0.23} & +0.15\\

\midrule
                        &  & \#2       & 81.93 & 73.86 & 81.61 & 0.83 & +0.07\\
\multirow{2}{*}{Google}                        &  & \#3   & \textbf{85.45} & \textbf{77.55} & \textbf{84.83} & \textbf{0.86} & +0.46\\
                        &  & \#3-1 & 83.91 & 75.56 & 83.51 & 0.85 & +0.82\\ 
 &  & \#3-2 & 85.18 & 77.07 & 84.55 & \textbf{0.86} & -0.65\\

\cmidrule{3-8}
&      & \#2       & 72.37 & 60.34 & 71.94 & 0.71 & -12.90\\
 \multirow{2}{*}{Broad}                       &  & \#3  & \textbf{82.25} & \textbf{69.63} & 80.62 & \textbf{0.80} & -3.38\\
                        &  & \#3-1 & 81.85 & 69.38 & \textbf{81.25} & \textbf{0.80} & -0.67\\ 
  & \multirow{2}{*}{1\%} & \#3-2 & 81.17 & 68.81 & 79.91 & 0.79 & -5.12\\

\cmidrule{3-8}
                        &  & \#2       & 64.15 & 52.32 & 63.34 & 0.63 & -19.47\\
  \multirow{2}{*}{BNC}                       &  & \#3   & 76.49 & 63.46 & 74.73 & 0.75 & -6.64\\
                        &  & \#3-1 & \textbf{77.50} & \textbf{64.64} & \textbf{76.65} & \textbf{0.77} & -1.77\\ 
   &  & \#3-2 & 74.27 & 61.72 & 72.73 & 0.73 & -8.34\\

\cmidrule{3-8}
                        &  & \#2       & \textbf{27.34} & \textbf{9.05} & \textbf{24.36} & \textbf{0.24} & -0.22\\
 \multirow{2}{*}{DUC}                       &  & \#3   & 26.63 & 8.57 & 23.93 & 0.23 & +1.73\\
                        &  & \#3-1 & 25.84 & 8.54 & 23.14 & 0.22 & -1.32\\
    &  & \#3-2 & 26.14 & 8.16 & 23.38 & 0.23 & -0.48\\

\midrule

                        &  & \#2       & 84.99 & 77.43 & 84.69 & 0.86 & +1.45\\
\multirow{2}{*}{Google} &  & \#3  & \textbf{86.88} & 79.55 & 86.26 & \textbf{0.87} & -0.16\\
                        &  & \#3-1 & 85.20 & 77.46 & 84.72 & 0.86 & +0.76\\ 
                        &  & \#3-2 & 86.80 & \textbf{79.58} & \textbf{86.29} & \textbf{0.87} & +0.12\\

\cmidrule{3-8}
&    & \#2         & 80.34 & 67.77 & 79.81 & 0.78 & -1.02\\
\multirow{2}{*}{Broad}                        &  & \#3  & 82.63 & 69.76 & 81.16 & \textbf{0.81} & -1.38\\
                        &  & \#3-1 & \textbf{82.80} & \textbf{70.39} & \textbf{82.05} & \textbf{0.81} & +0.90\\ 
  & \multirow{2}{*}{5\%} & \#3-2 & 82.66 & 69.81 & 81.16 & \textbf{0.81} & -1.08\\

\cmidrule{3-8}
                        &  & \#2       & 73.74 & 61.52 & 72.92 & 0.72 & -5.50\\
\multirow{2}{*}{BNC}                         &  & \#3  & 77.54 & 64.38 & 76.00 & 0.76 & -4.13\\
                        &  & \#3-1 & \textbf{77.62} & \textbf{64.58} & \textbf{76.45} & \textbf{0.77} & -1.49\\ 
   &  & \#3-2 & 77.40 & 64.20 & 75.81 & 0.76 & -4.03\\

\cmidrule{3-8}
                        &  & \#2       & \textbf{27.20} & \textbf{8.98} & \textbf{24.27} & \textbf{0.24} & +0.47\\
\multirow{2}{*}{DUC}                        &  & \#3   & 26.83 & 8.57 & 23.96 & 0.23 & +0.78\\
                        &  & \#3-1 & 26.25 & 8.27 & 23.49 & 0.23 & -1.22\\
    &  & \#3-2 & 26.46 & 8.31 & 23.62 & 0.23 & +1.32\\

\bottomrule
\end{tabular}

\end{adjustbox}

\caption{Ablation study of \quotes{length priming} for different training dataset sizes on \texttt{Google}, \texttt{Broad}, \texttt{BNC}, and \texttt{DUC}.}
\label{tab:qloradata2}
\end{table}

\begin{table}[t!]
\begin{adjustbox}{width=1\columnwidth,center}
\renewcommand{\arraystretch}{1.2}
\centering
\small

\begin{tabular}{cccccccc}

\toprule
\textbf{Data} & \textbf{Model} & \textbf{Instruction} & \textbf{R-1} & \textbf{R-2} & \textbf{R-L} & \textbf{F$_1$} & \textbf{$\Delta$ \textit{CR}}\\
\midrule

 &   \multirow{3}{*}{T5-XXL}                 & \#1              & 58.43 & 49.61 & 58.42 & 0.59 & +55.16 \\
  &                    & \#2          & 58.33 & 49.50 & 58.31 & 0.59 & +54.70  \\
\multirow{2}{*}{Google}       &                    & \#3         & 58.51 & 49.69 & 58.49 & 0.59 & +54.67 \\
\cmidrule{3-8}
       &   \multirow{3}{*}{UL2}              & \#1              & 59.17 & 50.33 & 59.03 & 0.60 & +51.05 \\
              &                 & \#2             & 58.59 & 49.68 & 58.52 & 0.59 & +53.35  \\
       &                 & \#3         & 60.59 & 51.66 & 60.53 & 0.61 & +49.27 \\
\midrule

 & \multirow{3}{*}{T5-XXL} &                  \#1 & 85.11 & 72.65 & 85.11 & 0.85 & +23.35 \\
  &  &                  \#2 & 85.08  &  72.56  & 85.08  & 0.85  & +22.85  \\
 
\multirow{2}{*}{Broad}       &     & \#3         & 85.22 & 72.77 & 85.21 & 0.85 & +22.94 \\
\cmidrule{3-8}
       &    \multirow{3}{*}{UL2}             & \#1              & 84.27 & 70.97 & 83.56 & 0.84 & +17.59 \\
              &                 & \#2              & 83.84 & 70.78 & 83.76 & 0.84 &+21.16 \\
       &                 & \#3         & 83.99 & 70.92 & 83.84 & 0.84 & +19.95 \\
\midrule

    &  \multirow{3}{*}{T5-XXL}                  & \#1              & 81.43 & 68.46 & 81.43 & 0.82 & +27.06 \\
        &         & \#2              &  81.39 & 68.49  & 81.39  & 0.82  & +27.18  \\
\multirow{2}{*}{BNC}                          &          & \#3 & 81.42 & 68.51 & 81.40 & 0.81 & +26.97 \\
\cmidrule{3-8}
&\multirow{3}{*}{UL2} & \#1 & 80.20 & 67.07 & 79.00 & 0.80 & +19.77 \\
         &            & \#2 & 80.29   & 67.01   & 80.03   & 0.80   &  +24.12  \\
                   &  & \#3 & 79.81 & 66.61 & 79.32 & 0.80 & +21.92 \\



\bottomrule
\end{tabular}
\end{adjustbox}
\caption{Experimental results for the zero-shot instruction-based FLAN models using instruction templates in Table~\ref{tab:instruction_format}.}
\label{tab:flan1}
\end{table}

\begin{table*}[t]
\begin{adjustbox}{width=1.9\columnwidth,center}
\centering
\small
\renewcommand{\arraystretch}{1.2}
\begin{tabular}{ccl}
\toprule
\# &\textbf{Constraint} & \textbf{Instruction}\\
\midrule
1 & \xmark             & Sentence:\textbackslash n\{src\}\textbackslash nSummarize without the less important words would be:\textbackslash n\\
\cmidrule{3-3}
2 & Length w/o priming & Sentence:\textbackslash n\{src\}\textbackslash nSummarize without the less important \{del\} words would be:\textbackslash n\\
\cmidrule{3-3}
\multirow{2}{*}{3} & \multirow{2}{*}{Length}  & Sentence with \{src len\} words:\textbackslash n\{src\}\textbackslash nSummarize in \{keep\} words without the \\
&&less important \{del\} words would be:\textbackslash n \\
\bottomrule
\end{tabular}
\end{adjustbox}
\caption{Modified instruction templates for the FLAN models.}

\label{tab:instruction_format2}
\end{table*}

\end{document}